\documentclass[a4paper,fleqn]{cas-dc}

\usepackage[numbers]{natbib}
\usepackage{threeparttable}
\usepackage{multirow}
\usepackage{algorithm}
\usepackage{algorithmic}

\begin{document}
\shorttitle{RoNFA: Robust Neural Field-based Approach for Few-Shot Image Classification with Noisy Labels}
\shortauthors{Nan Xiang et~al.}

\title [mode = title]{RoNFA: Robust Neural Field-based Approach for Few-Shot Image Classification with Noisy Labels}                      

%

\author[1]{Nan Xiang}
\ead{2306301044@st.gxu.edu.cn}

\credit{ Methodology,Conceptualization, Investigation, Writing-Review \& Editing.}

\author[1]{Lifeng Xing}
\ead{2406301052@st.gxu.edu.cn}

\credit{ Validation.}

\author[1,2]{Dequan Jin\corref{cor1}}
\ead{dqjin@gxu.edu.cn}
\cortext[cor1]{Corresponding author}
\credit{ Project administration, Supervision, Resources, Funding acquisition.}

\affiliation[1]{organization={School of Mathematics and Information Science},
	addressline={No.100 East Daxue Road}, 
	city={Nanning},
	postcode={530004}, 
	state={Guangxi},
	country={China}}

\affiliation[2]{organization={Center for applied mathematics of Guangxi, Guangxi University},
	addressline={No.100 East Daxue Road}, 
	city={Nanning},
	postcode={530004}, 
	state={Guangxi},
	country={China}}

%

\begin{abstract}
In few-shot learning (FSL), the labeled samples are scarce. Thus, label errors can significantly reduce classification accuracy. Since label errors are inevitable in realistic learning tasks, improving the robustness of the model in the presence of label errors is critical. This paper proposes a new robust neural field-based image approach (RoNFA) for few-shot image classification with noisy labels. RoNFA consists of two neural fields for feature and category representation. They correspond to the feature space and category set. Each neuron in the field for category representation (FCR) has a receptive field (RF) on the field for feature representation (FFR) centered at the representative neuron for its category generated by soft clustering. In the prediction stage, the range of these receptive fields adapts according to the neuronal activation in FCR to ensure prediction accuracy. These learning strategies provide the proposed model with excellent few-shot learning capability and strong robustness against label noises. The experimental results on real-world FSL datasets with three different types of label noise demonstrate that the proposed method significantly outperforms state-of-the-art FSL methods. Its accuracy obtained in the presence of noisy labels even surpasses the results obtained by state-of-the-art FSL methods trained on clean support sets, indicating its strong robustness against noisy labels.
\end{abstract}

%


\maketitle

\section{Introduction}
Few-shot learning (FSL) methods aim to train classifiers for new categories using only a few labeled samples. Most discussions on FSL, often assume that the support set samples are accurately labeled to represent their categories. However, this assumption rarely holds in real-world scenarios since error labels can happen in sample collecting, labeling, or their transition, almost inevitable due to weakly supervised annotation methods, ambiguities, or human errors \cite{efedddcd838340f08db5e56563ecdcf3,Tsipras2020FromIT,northcutt2021pervasive}. The performance of typical FSL models significantly relies on the accurately labeled sample. When trained on the samples with noisy labels, many FSL methods may dramatically lose their accuracy and struggle in practical applications \cite{10.1145/3446776}. 

Most current learning methods for handling mislabeled samples are large-sample methods \cite{9880080,Li2020DivideMix:,10.5555/3495724.3497431}. These methods assume that the majority of the dataset is correctly labeled, enabling the model to statistically identify and mitigate the impact of a small number of mislabeled samples by leveraging correctly labeled samples to estimate the noise distribution or correct erroneous labels \cite{SUN2022108467}. For instance, noise-tolerant loss functions such as the mean absolute error or robust cross-entropy focus less on outliers, reducing their influence during optimization \cite{9156369,Li2020DivideMix:}. Alternatively, label correction approaches may iteratively identify and relabel noisy samples by comparing their predictions with model confidence scores \cite{9710363}. However,  extensive noise may lead to unreliable data distribution when the sample is small. Thus, the denoising effectiveness significantly diminishes when applied to small sample datasets where mislabeled samples constitute a large proportion of the data. 


We address FSL with noisy labels and propose a new visual cognition-inspired model (VCIM). We present a new classifier architecture consisting of two neural fields for sample feature and category representation. The connections between the two fields follow the receptive field theory and Hebbian rules, and its prediction utilizes some efficient learning strategies inspired by visual cognitive behaviors. To achieve efficiency and robustness in classification, we employ soft K-means clustering to optimize the class distribution and add it to the proposed classifier, providing excellent few-shot learning performance in the presence of noisy labels. We extensively evaluate our proposed method on real-world FSL datasets with three different types of label noise and compare it with state-of-the-art methods. The experimental results indicate that the proposed model achieves superior results to these methods, demonstrating its dramatic performance and strong robustness against label noises in FSL. The main contributions of this paper are as follows:
\begin{itemize}
	\item We introduce a robust neural field-base architecture for few-shot learning against noisy labels. 
	\item We present some learning strategies for model training and predicting, effectively improving the computation efficiency and mode robustness. We introduce a local learning mechanism based on the Hebbian rule, enabling the model to operate without relying on multi-layer backpropagation of error signals.
	
	\item Experiments on two real-world datasets demonstrate that our method achieves state-of-the-art accuracy on real-world datasets without fine-tuning, exhibits superior robustness and performance compared to existing algorithms in FSL with noisy labels. 
\end{itemize}

\section{Related Works}
\label{sec:rw}
\subsection{Few-Shot Learning}
Transfer learning dramatically improves the FSL performance of deep neural networks. Prototype Networks average support features to create class prototypes and predict query classes via nearest neighbors \cite{NIPS2017_cb8da676}. SNAIL combines temporal convolution and soft attention for meta-learning \cite{Mishra2017ASN}. MetaQDA integrates Bayesian meta-learning with shallow learning to handle data scarcity, class imbalance, and uncertainty \cite{Zhang2021ShallowBM}. FewTRUE encodes input patches to establish semantic correspondences between localized regions, using meta-tuned encoders and marker reweighting to avoid supervisory collapse \cite{10.5555/3600270.3600529}.  

Efficient feature extraction techniques can significantly improve the FSL performance. HCTransformers use hierarchical cascade transformers with spectral pooling to reduce foreground-background ambiguity and optimize parameters via latent attributes \cite{9879324}. GPICL leverages Transformers as general-purpose context learners, improving generalization by mitigating memory constraints through biased training interventions \cite{kirsch2022generalpurpose}. CAML learns new visual concepts during inference without fine-tuning, mimicking large-scale language models \cite{fifty2024contextaware}. BPA enhances FSL by encoding higher-order feature relationships to optimize tasks like feature matching and grouping \cite{shalam2024the}.

\subsection{Noisy Labels}
Some approaches for FSL with noisy labels focus on designing robust loss functions. Peer loss function learns from noisy labels without prior knowledge of the noise rate \cite{10.5555/3524938.3525516}. The active-passive loss framework combines two robust loss functions to enhance noise resistance \cite{Ma2020NormalizedLF}. 

Recent works have developed some effective label correction strategies. ProSelfLC updates labels via self-prediction of model outputs\cite{9578233}, while a meta-learning approach estimates soft labels through meta-gradient descent using noiseless metadata to avoid manual hyperparameter tuning \cite{Wu_Shu_Xie_Zhao_Meng_2021}. MLC treats label correction as a meta-process, employing a correction network to generate optimized labels jointly with the primary model \cite{article}. SNSCL focuses on representation distinguishability by designing a noise-tolerant supervised contrastive loss, incorporating weight-aware mechanisms for label correction, and optimizing momentum queue lists for further improvement on representation \cite{10205042}.

The FSL with noisy labels is more challenging and thus rarely discussed. RNNP refines class prototypes by generating hybrid features from the support examples of each class to improve query image classification\cite{9423328}. TraNFS improves upon the prototype used by ProtoNet and utilizes the Transformer's attention mechanism to weigh mislabeled versus correctly labeled samples \cite{Liang2022}.

\section{Preliminaries}\label{sec:pre}
The FSL task aims to create an effective way to pre-train a classifier on the base classes in $C^b$ with sufficient labeled samples and predict new classes in $C^n$ with a few labeled samples where $C^n$ does not share any common classes with $C^b$, i.e., $C^b \cap C^n = \emptyset$. FSL classification tasks are typically N-way K-shot, where N is the number of classes in $C^n$, and K is the number of labeled samples per class. The support set is denoted as $S = \{ x_1^1, x_2^1, \dots, x_K^N \}$. The query set $Q = \{ x_1^*, x_2^*, \dots \}$ consists of unlabeled samples of the N classes. 

FSL models leverage transfer learning and meta-learning frameworks. During training, the model learns generic feature embeddings from the base classes and transfers their features to new tasks. Since K is usually very small, noisy support samples significantly impact the model performance, undermining feature reliability, causing incorrect class representation, and making the prediction challenging. 

\section{Methodology}\label{sec:method}


Suppose $\mathbf{I}_i$, $i=1,2,\cdots,N$ are image samples of $m$ categories in the support set $S$. Denote an image vector or matrix of the $c$th category by $\mathbf{I}_i^c$, $i=1,2,\cdots, N_c$. $N_c$ is the number of its support samples. Let \(\mathbf{x}_i = Net(\mathbf{I}_i)\), $i=1,2,\cdots,N$, and \(\mathbf{x}_j^c = Net(\mathbf{I}_i^c)\), $j=1,2,\cdots, N_c$ be the extracted feature vector, where $Net(\cdot)$ is a deep neural network performing as an feature extractor. Our proposed modeling framework is shown in Figure \ref{mod}.
\subsection{Representative for Category}
\begin{figure*}
	\centering
	\includegraphics[width=4in]{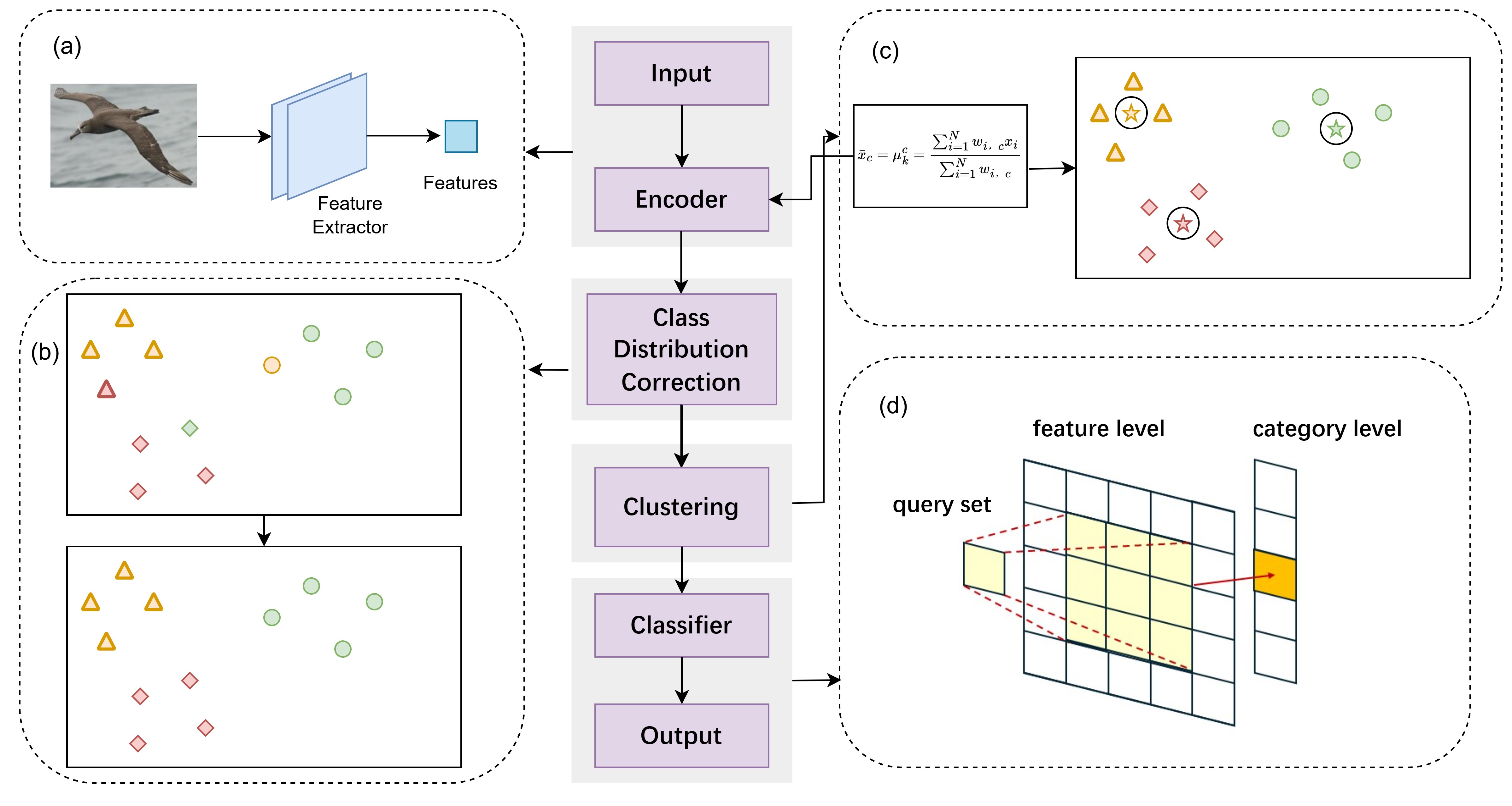}
	\caption{(a) The model extracts features from the input image through the feature extractor. (b) Correcting the class distribution of label-noise samples using soft K-means clustering. (c) Calculating class representative samples after clustering using a weighted mean. (d) Classifying query set samples based on visual cognitive mechanisms and using scale adaptation.}\label{mod}
\end{figure*}

The support set often contains very few samples and practical scenarios frequently introduce noisy labels. Label noises pose a significant challenge since traditional classifiers rely on accurate labels and assume that they reflect the correct class distribution, but noisy labels violate this. They cause the training to be under an incorrect sample distribution and significantly degrade model performance on the query set. To address these issues, a feasible way is correcting the label errors with sample distribution, as shown in Figure \ref{mod}. Following this idea, we cluster the support samples to generate the representatives for their categories according to the clustering results. Since we have known the number of categories, we leverage K-means clustering and let $K$ be category number $m$. 




The K-means clustering is an iterative process. It usually randomly selects $K$ initial cluster centers before it starts. However, this eliminates the connection between the obtained cluster and the sample category. For this issue, we calculate the center of the support samples in the $c$th category $$\mathbf{\mu}_0^{c}=\frac{1}{N_c}\sum_{j=1}^{N_{c}} \mathbf{x}_j^{c}$$  and employ it as the initial center for the $c$th cluster instead to keep their correspondence. After that, we calculate the Euclidean distance from the $i$th support sample to the $c$th class center by $$d^0_{i,{c}} = \| \mathbf{x}_i - \mathbf{\mu}_{0}^{c} \|_2^2.$$ 

Since the number of support samples is limited, the clustering results are easily affected by randomness in sample selection. To reduce its impact, we use a soft strategy that allows samples to belong to multiple clusters rather than rigidly assigning each data point to a single cluster. To find the new center of the $c$th cluster at the $k$th iterative step, $k=0,1,2,\cdots$, we first calculate the soft-assignment weight with a Gaussian kernel by the following equation: $$w_{i,c} = \frac{e^{-d_{i,c}^k}}{\sum_{k=1}^{m} e^{-d_{i,k}^k}}, $$ where $i=1,2,\cdots, N$ and $c=1,2,\cdots,m$. The weight $w_{i,c}$ indicates the probability for the sample $\mathbf{x}_i$ belonging to the class $c$. Then we update the $c$th cluster center $\mu_{k}^{c}$ by the following equation:

\begin{eqnarray}\label{sys:clu}	
	\mu_k^{c} = \frac{\sum_{i=1}^{N} w_{i,c} \mathbf{x}_i}{\sum_{i=1}^{N} w_{i,c}}, c=1,2,\cdots, m.
\end{eqnarray}
The obtained cluster centers are weighted averages, which better describe the realistic sample distribution and are less impacted by the randomness in sample selection.

We further calculate the Euclidean distance from the $i$th support sample to the obtain $c$th cluster center by 
\begin{eqnarray}
	d^{k}_{i,{c}} = \| \mathbf{x}_i - \mathbf{\mu}_{k}^{c} \|_2^2.
\end{eqnarray}
Repeat the clustering process. When the $\mathbf{\mu}_{k}^{c}$, $c=1,2,\cdots,m$ become stable, that is, $$\sum_{c=1}^m|\mathbf{\mu}_{k}^{c}-\mu_{k-1}^{c}|<\epsilon,$$ where $\epsilon$ is a small positive constant, or $k$ reaches a given upper bound $k_{up}$, we end the clustering process and let the primary representative sample of the $c$th category $$\bar{x}_c=\mathbf{\mu}_{k}^{c}, c=1,2,\cdots,m.$$ The K-means clustering is an unsupervised process insensitive to label noises. The soft strategy reduces the impact of the significant randomness in sample selection in few-shot learning. They together ensure the representatives of $\bar{\mathbf{x}}_c$ for its category. Since there is only one representative for each category, in these procedure, we may not relabel some support samples because of their low weights. In this case, we have to abandon them to reduce the impact of label error. The fewer support samples requires the model more powerful few-shot learning capability. 
\subsection{Framework of Classifier}

The learning process of most current neural network models relies on the error backpropagation (BP) algorithm, which iteratively adjusts the network's connection weights layer by layer based on the difference between the network's output and the expected output. BP algorithm gives neural network models powerful learning capabilities but also results in relatively slow weight adjustment speeds, requiring a large number of training samples, so requiring a large sample for training. Although transfer learning strategies can mitigate this issue through pre-training and fine-tuning, the network's response to very few samples is still slow and prone to overfitting. Thus, we will try a way not to use the typical forward network's topology or the BP strategy. 

The proposed classifier utilize two neural fields, one for feature representation and the other one for category representation. The field for feature representation (FFR) consists of $m$ neurons $v_c$ located at $\bar{\mathbf{x}}_c$, $c=1,2,\cdots,m$ and altering according to support samples. The field for category representation (FCR) also consists of $m$ neurons $u_c$, $c=1,2,\cdots,m$. They receive stimuli from their receptive fields in FFR. 

Recent research on neuroscience discovered that the receptive fields drift during learning, changing the sensitivity to its inputs \cite{Shine2021,Qin2023}. Inspired from this, we suppose the receptive field of an FCR neuron centered at corresponding $v_c$ whose position relies on the support samples. For a new input image sample $\mathbf{I}$, let $\mathbf{x}=Net(\mathbf{I})$. Then its impact on the FCR neurons are formulated by the following equation:
\begin{eqnarray}\label{fun:phi}
	\phi_\sigma(\mathbf{x},\bar{\mathbf{x}}_c) =Ae^{ -\frac{1}{2} \frac{\|\mathbf{x}-\bar{\mathbf{x}}_c\|_2^2}{\sigma^2} } - Be^{ -\frac{1}{2} \frac{\|\mathbf{x}-\bar{\mathbf{x}}_c\|_2^2}{(3\sigma)^2} },
\end{eqnarray}
whose right-hand side is the difference of Gaussian functions determining the shape of the receptive field, homogeneous with a Mexican hat shape. The constant $\sigma>0$ determines the excitatory and inhibitory ranges of $\phi_\sigma$. Denote $\phi_\sigma(r)=\phi_\sigma(\mathbf{x},\mathbf{x}')$ by letting $r=\frac{1}{2}\|\mathbf{x}-\bar{\mathbf{x}}_c\|_2$, then the equation $\phi_\sigma(r)=0$ has only one real solution $r=\frac{3\sqrt{\ln{3}}}{2}\sigma$. Generally, the constants $A=\frac{1}{\sqrt{2\pi}\sigma}$ and $A=\frac{1}{3\sqrt{2\pi}\sigma}$ when consider the Gaussian function as probability density, but leading to the difficulty in discussing the excitatory and inhibitory radius of receptive field. Therefore, we simplify their selection by letting $A=1.5$ and $B=0.5$.

The response of the FCR neuron $u_c$ to the input stimulus is determined by: 
\begin{eqnarray}\label{sys:c-p}
	u_c = \varphi \left( \phi_\sigma(\mathbf{x},\bar{\mathbf{x}}_c) - h_u \right),
\end{eqnarray}
where $c=1,2,\cdots,m$.  $h_u>0$ is the resting level, ensuring any input weaker than it cannot activate the neuron. The function $\varphi(\cdot)$ is a nonlinear activation function to characterize the activation of the neuron, defined by the following equation:
\begin{eqnarray}\label{sys:act}
	\varphi(u) =
	\begin{cases}
		1 - \exp(-u), & \text{if } u \geq 0 \\
		0, & \text{if } u \leq 0
	\end{cases}.
\end{eqnarray}
When the input stimulus strength $u$ exceeds the resting level, the neuron is activated quickly, whereas when the stimulus strength is below the resting level, the neuron remains inactivated. 

When a neuron in the FCR responds positively to a stimulus $v_i$, i.e., $u_c > 0$, the corresponding category can be considered a potential candidate for the input sample $v_i$. This mechanism enables the model to effectively match input features to category representations despite multi-category competition, modeling the selective responses of neurons. However, this response highly depends on the receptive field size determined by $\sigma$, which directly influences the sensitivity and activation strength. Different values of $\sigma$ may result in different activation distributions for the same input, requiring a balance between accuracy and generalization. 

\subsection{Scale Adaptation}
In the prediction stage, the number of activated neurons in the FCR can present three possible scenarios depending on the receptive field size: no neuron activated, single neuron activated, or multiple neurons activated. The ideal scenario is the second one, as it indicates the category of the input stimulus. If the receptive field is too small or too large, the first or third scenarios may occur, suggesting the model cannot identify a specific category for the input stimulus.\par
To address this problem, we introduce a simple strategy as follows: 
\begin{enumerate}
	\item Initialize the scale $\sigma_0$, its upper bound $\sigma_{max}$ and lower bound $ \sigma_{\text{min}}$, $\sigma_{max} = \sigma_{\text{min}} = 0$, and the  tuning ratio parameter $\lambda$. 
	\item Calculate the response of the neurons in the FCR. If the number of activated neurons $n_0 > 1$, indicating that the receptive field is too large. Let $\sigma_{max} = \sigma_{k-1}$. Update $\sigma_k$ by letting $\sigma_k = \sigma_{max} - \lambda (\sigma_{\text{max}} - \sigma_{\text{min}})$.
	\item If $n_0 = 0$, when $\sigma_{max} = 0$, let $\sigma_{k} = \sigma_{k-1}/\lambda$; when $\sigma_{max} \neq 0$, let $\sigma_{min} = \sigma_{k-1}$ and $\sigma_k = \sigma_{max} - \lambda (\sigma_{\text{max}} - \sigma_{\text{min}})$.
\end{enumerate}
By iterating this process repeatedly, we gradually adjust the $\sigma$ based on the number of activated neurons in each trial until the stimulus activates exactly one neuron in the FCR. In other words, the receptive field parameter $\sigma$ is adaptively optimized based on the number of activated neurons, ensuring the stability and accuracy of the classification results.
	
	
	\begin{table*}[ht]
		\centering
		\caption{Performance of FSL experiments with Symmetric label swap noise and paired label swap noise on MiniImageNet dataset}\label{tab:mini}
		\resizebox{0.5\textwidth}{!}{
			\begin{tabular}{l|ccccccc}
				\hline
				Model \textbackslash Noise Proportion & Backbone & 0\% & 20\%sym & 40\%sym & 60\%sym & 40\%pair \\
				\hline
				Matching Networks\footnotemark[1] & Conv4 & 62.16$\pm$0.17 & 56.21$\pm$0.18 & 46.18$\pm$0.18 & 34.66$\pm$0.18 & 43.53$\pm$0.17 \\
				Vanilla ProtoNet\footnotemark[1] & Conv4 & 68.27$\pm$0.16 & 62.43$\pm$0.17 & 51.41$\pm$0.19 & 38.33$\pm$0.19 & 47.77$\pm$0.19 \\
				TraNFS-3\footnotemark[1] & Conv4 & 68.53$\pm$0.17 & 65.08$\pm$0.18 & 56.65$\pm$0.21 & 42.60$\pm$0.24 & 53.96$\pm$0.23 \\
				RNNP\footnotemark[1] & Conv4 & 68.38$\pm$0.16 & 62.43$\pm$0.17 & 51.62$\pm$0.19 & 38.45$\pm$0.19 & 47.88$\pm$0.19 \\
				Vanilla ProtoNet & VIT & 98.46$\pm$0.01 & 97.59$\pm$0.02 & 96.39$\pm$0.03 & \underline{88.27$\pm$0.09} & \underline{91.07$\pm$0.08} \\
				RNNP & VIT & \underline{98.57$\pm$0.01} & \underline{98.20$\pm$0.02} & \underline{96.87$\pm$0.07} & 77.34$\pm$0.24 & 88.04$\pm$0.19 \\
				VCIM(ours) & VIT & \textbf{99.17$\pm$0.01} & \textbf{99.12$\pm$0.01} & \textbf{99.11$\pm$0.01} &\textbf{98.33$\pm$0.05} &
				\textbf{{98.76$\pm$0.03}} \\
				\hline
			\end{tabular}
		}
		\begin{tablenotes}
			
			\footnotesize
			\item \textit{The results in Tables, \footnotemark[1] by \cite{Liang2022}. 
			}
		\end{tablenotes}
	\end{table*}

	\section{Experiments}
	\label{sec:er}
	\subsection{Experimental Setup}
	\subsubsection{Datasets} We conduct experiments on two FSL datasets: MiniImageNet \cite{10.5555/3157382.3157504} and TieredImageNet \cite{ren2018metalearning}. Both MiniImageNet and TieredImageNet consist of 84×84 pixel images. MiniImageNet contains  64 classes for training, 16 for validation, and 20 for testing classes for training, with 60,000 images in total. TieredImageNet has 351 classes for training, 97 for validation, and 160 for testing, with 779,165 images in total. \par
	\subsubsection{Label Noise Types}  We explore the following three forms of labeling noise:\par
	Symmetric label swap noise refers to the type of noise described in \cite{10.5555/2969239.2969241}. Mislabel samples are randomly and uniformly selected from the other categories in the current task to ensure that they differ from and do not exceed the number of original clean categories.\par
	Paired label swap noise described in \cite{10.5555/3327757.3327944} is a more challenging type of noise. Each category is consistently assigned a fixed mislabeled category, simulating real-world labeling errors where some categories are easily confused. In the experiments, we randomly assigned noise categories for each task.\par
	
	Outlier noise refers to samples originating from classes outside the current task class \cite{Liang2022}. For this,  images selected from 350 non-MiniImageNet and non-TieredImageNet classes of ImageNet as noises ensure that the outlier noise samples in the meta-testing set come from classes the model has not encountered.\par
	The noise proportion in the support set is the percentage of the total sample count. We focus on noise levels that allow clean categories to remain identifiable under reasonable conditions. 
	
	The proportion of pairwise label-swapping noise is 40\% since it is selected in the same way as the Symmetric label swap noise in 5-way 5-shot tasks when the noise proportion is 60\% and a higher proportion would obscure clean categories or reduce them to a minority, making performance evaluation unreliable. 
	\begin{table*}[ht]
		\centering
		\caption{Performance of FSL experiments with Symmetric label swap noise and paired label swap noise on TieredImageNet dataset}\label{tab:tiered}
		\resizebox{0.5\textwidth}{!}{
			\begin{tabular}{l|cccccc}
				\hline
				Model \textbackslash Noise Proportion & Backbone & 0\% & 20\%sym & 40\%sym & 60\%sym & 40\%pair \\
				\hline
				Matching Networks\footnotemark[1] & Conv4 & 64.92$\pm$0.19 & 59.2$\pm$0.20 & 49.12$\pm$0.20 & 36.8$\pm$0.19 & 46.13$\pm$0.19 \\
				Vanilla ProtoNet\footnotemark[1] & Conv4 & 71.36$\pm$0.18 & 66.15$\pm$0.19 & 55.05$\pm$0.21 & 40.61$\pm$0.21 & 50.85$\pm$0.21 \\
				TraNFS-3\footnotemark[1] & Conv4 & 71.17$\pm$0.19 & 67.67$\pm$0.20 & 58.88$\pm$0.23 & 44.21$\pm$0.25 & 55.12$\pm$0.24 \\
				RNNP\footnotemark[1] & Conv4 & 71.36$\pm$0.18 & 65.95$\pm$0.19 & 54.86$\pm$0.21 & 40.63$\pm$0.21 & 50.91$\pm$0.20 \\
				Vanilla ProtoNet & VIT & \underline{94.67$\pm$0.06} & 92.29$\pm$0.07 & \underline{89.38$\pm$0.08} & \underline{74.74$\pm$0.15} & \underline{82.35$\pm$0.12} \\
				RNNP & VIT & 94.42$\pm$0.06 & \underline{92.62$\pm$0.08} & 88.62$\pm$0.13 & 63.68$\pm$0.23 & 77.69$\pm$0.20 \\
				VCIM(ours) & VIT &  \textbf{95.88$\pm$0.05}& \textbf{95.49$\pm$0.06} & \textbf{94.85$\pm$0.08} & \textbf{90.57$\pm$0.15} & \textbf{93.82$\pm$0.11} \\
				\hline
			\end{tabular}
		}
		\begin{tablenotes}
			
			\footnotesize
			\item \textit{The results in Tables, \footnotemark[1] by \cite{Liang2022}. 
			}
		\end{tablenotes}
	\end{table*}

	\subsubsection{Implementation Details} We selected Vision Transformer (ViT) presented by \cite{dosovitskiy2021an} as a frozen feature encoder without fine-tuning. We conduct 600 tests on each dataset and evaluate the model's classification accuracy under different proportions of symmetric label swap swap noise, pairwise label swap noise, and outlier noise to demonstrate the model's adaptability and denoising performance in complex noise environments in comparison with state-of-the-art models in FSL with noisy labels. 
	
	\subsection{Noisy Few-Shot Results}
	\begin{figure}
		\centering
		\begin{minipage}[t]{0.2\textwidth}
			\centering
			\includegraphics[width=\textwidth]{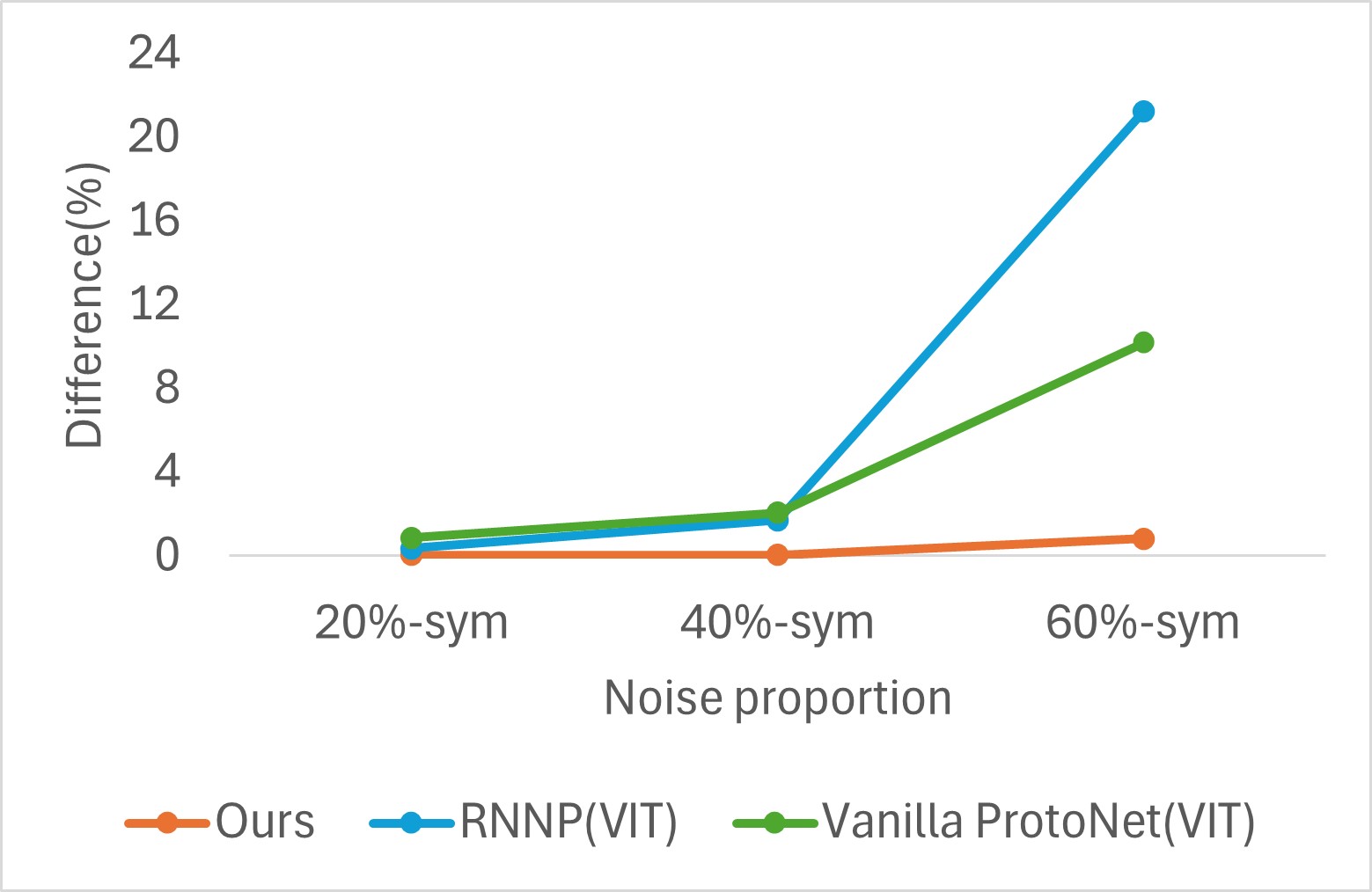}
			(a)
		\end{minipage}
		\begin{minipage}[t]{0.2\textwidth}
			\centering
			\includegraphics[width=\textwidth]{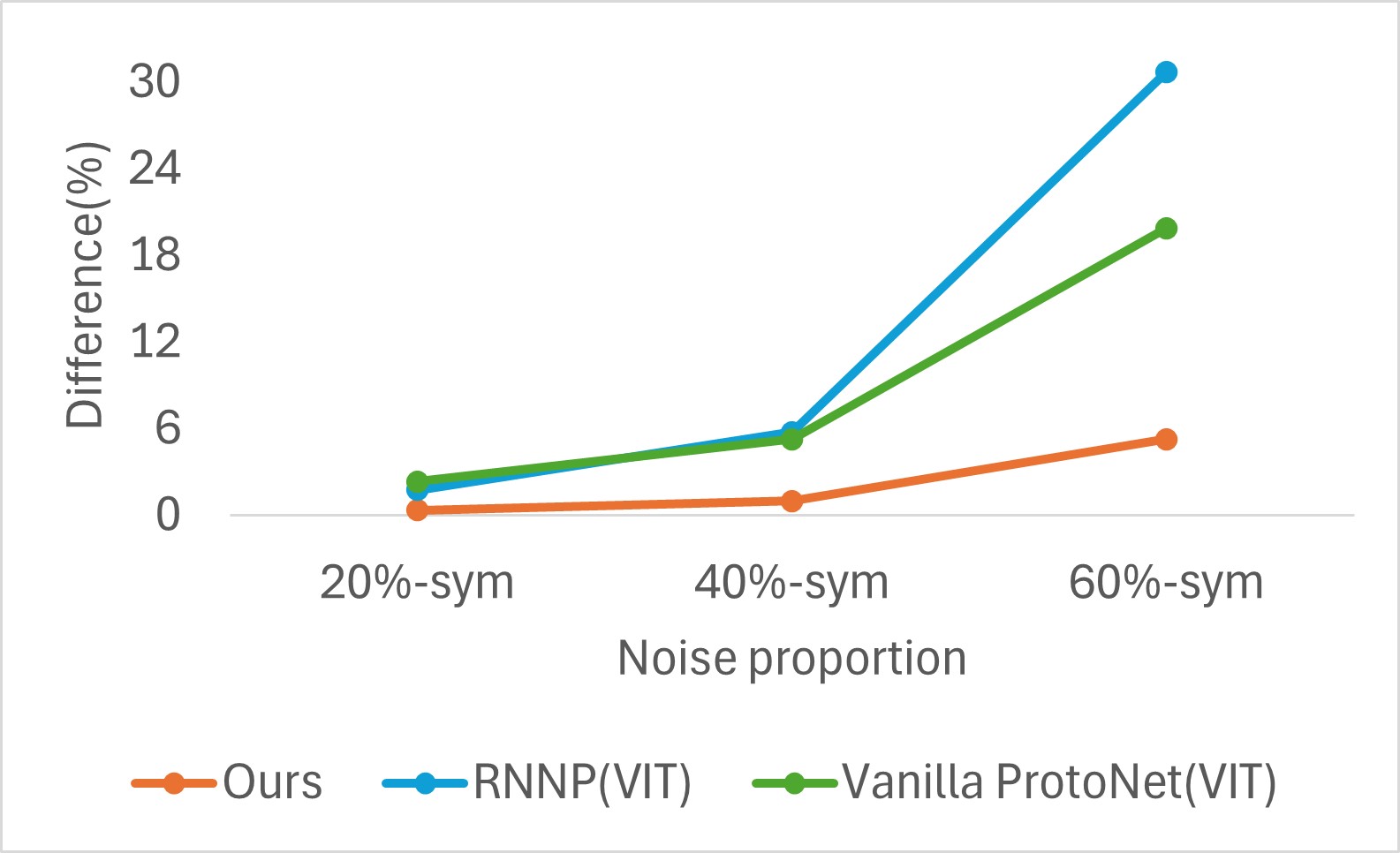}
			(b)
		\end{minipage}
		\begin{minipage}[t]{0.2\textwidth}
			\centering
			\includegraphics[width=\textwidth]{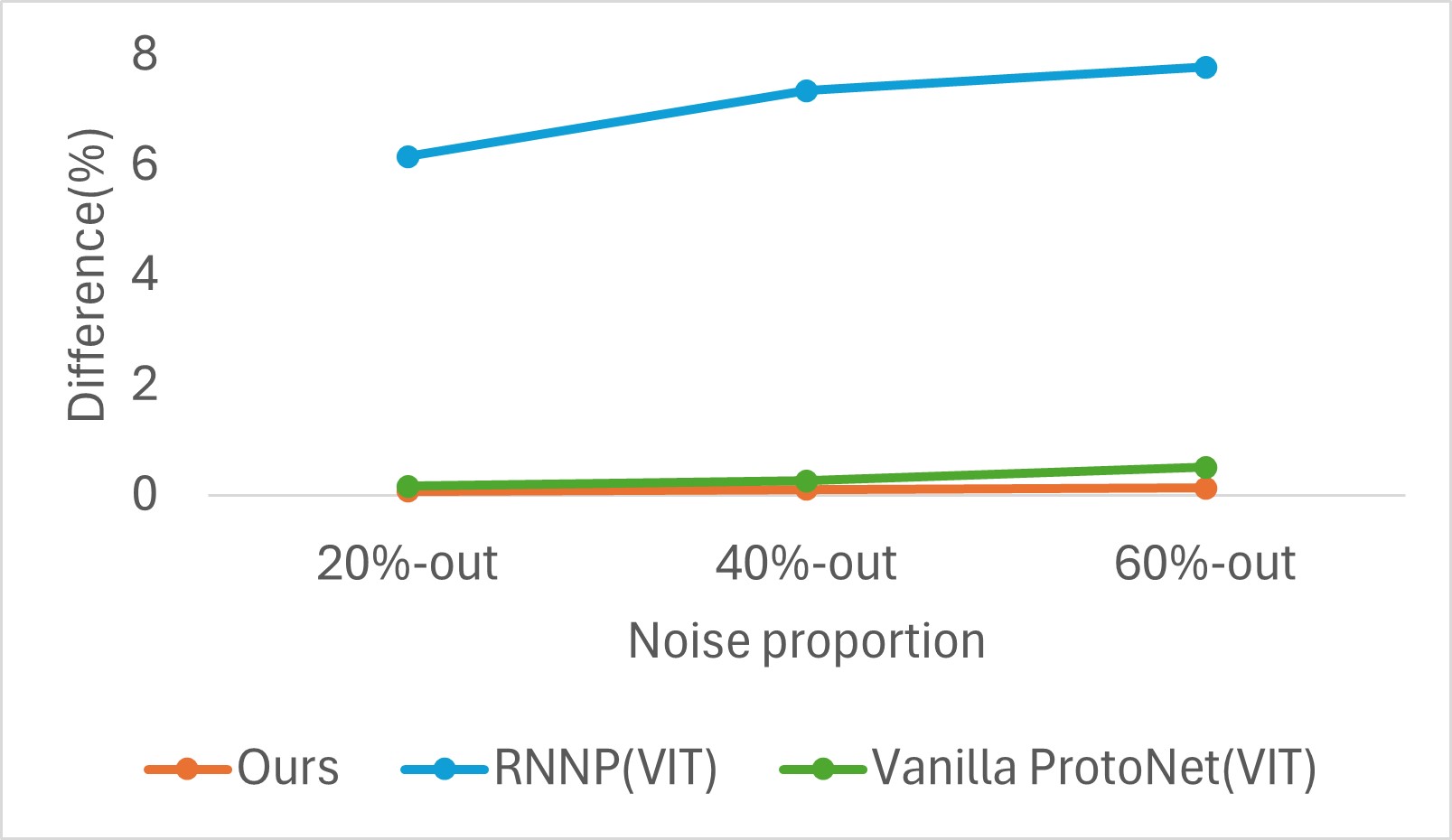}
			(c)
		\end{minipage}
		\begin{minipage}[t]{0.2\textwidth}
			\centering
			\includegraphics[width=\textwidth]{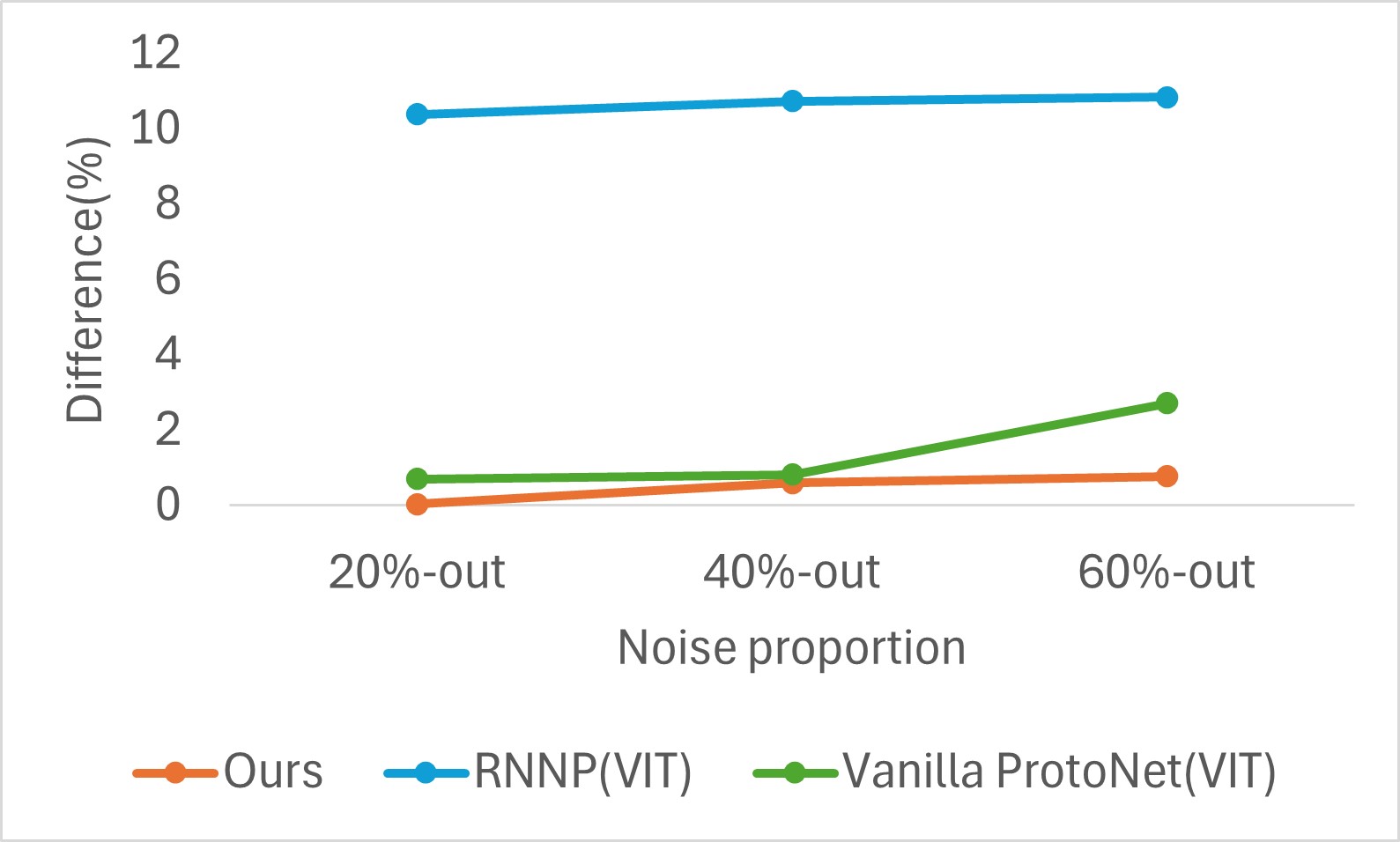}
			(d)
		\end{minipage}
		\caption{(a) The decline on accuracy with increasing symmetric label swap noise on MiniImageNet; (b) The decline on accuracy with increasing symmetric label swap noise on TieredImageNet;  (c) The decline on accuracy with increasing outlier noise on MiniImageNet; (d) The decline on accuracy with increasing outlier noise on TieredImageNet.}\label{acc}
	\end{figure}
	We test the proposed model on MiniImageNet with 0\% to 60\% symmetric and 40\% paired label swap noise. As shown in Table \ref{tab:mini}, the performance of all models degrades when the proportion of the paired label swap noise increases. However, the proposed model shows strong robustness against noise. Its accuracy drops less than 1.0\% when the noise increases from 0\% to 60\%, as shown in Figure \ref{acc}(a), while Vanilla ProtoNet(VIT) and RNNP(VIT) drop about  10\% and 21\%. Its accuracy with 60\% symmetric label swap noise is even comparable with the results obtained by Vanilla ProtoNet(VIT) and RNNP(VIT) without noise. The proposed model also performs well under the more challenging 40\% paired label swap noise condition, achieving 7.69\%  ahead of the second-best and reaching a dramatic accuracy of 98.76\%. 
	
	We test the proposed model on tieredImageNet with 0\% to 60\% symmetric and 40\% paired label swap noise, as shown in Table \ref{tab:tiered}. The proposed model still maintains superior performance compared to other models. It achieves an accuracy advantage of 15.83\% and 26.89\% over Vinilla ProtoNet(VIT) and RNNP(VIT) with 60\% symmetric label swap noise and 11.47\% and 16.13\% with 40\% pairwise label swap noise. The proposed model shows strong robustness against noise. Its accuracy drops no more than 6\% when the noise increases from 0\% to 60\%, as shown in Figure \ref{acc}(b), while Vanilla ProtoNet and RNNP drop about  20\% and 31\%.

	We also test the proposed model on MiniImageNet and tieredImageNet with 0\% to 60\% outlier label noise, as shown in Table \ref{tab:out1} and \ref{tab:out2}. It achieves an accuracy advantage of 1.09\% and 3.15\% over Vinilla ProtoNet(VIT) with 60\% symmetric label swap noise on the two datasets. The proposed model is dramatically robust to outlier noise. Its accuracy drops 0.13\% adn 0.78\% when the noise increases from 0\% to 60\%, as shown in Figure \ref{acc}(c) and (d), while Vanilla ProtoNet and RNNP drop about  0.51\% and 2.72\%. Its accuracy with 60\% symmetric label swap noise is even higher than the results obtained by Vanilla ProtoNet and RNNP without noise. 
	
	\begin{table*}[ht]
		\centering
		\caption{Performance of FSL experiments with outlier label noise on MiniImageNet dataset}\label{tab:out1}
		\resizebox{0.5\textwidth}{!}{
			\begin{tabular}{lccccccc}
				\hline
				Method & Backbone & 0\% & 20\% & 40\% & 60\%\\
				\hline
				Matching Networks\footnotemark[1] & Conv4 & 62.05$\pm$0.17 & 57.69$\pm$0.18 & 51.32$\pm$0.19 & 42.39$\pm$0.19 \\
				Vanilla ProtoNet\footnotemark[1] & Conv4 & 68.18$\pm$0.16 & 63.92$\pm$0.17 & 57.07$\pm$0.18 & 46.99$\pm$0.20 \\
				TraNFS-3\footnotemark[1] &Conv4 & 68.11$\pm$0.17 & 64.96$\pm$0.18 & 59.03$\pm$0.20 & 47.69$\pm$0.22 \\
				RNNP\footnotemark[1] & Conv4 & 68.17$\pm$0.16 & 63.80$\pm$0.17 & 56.97$\pm$0.18 & 46.92$\pm$0.20 \\
				Vanilla ProtoNet & VIT & 98.46$\pm$0.01 & \underline{98.30$\pm$0.02} & \underline{98.20$\pm$0.02} & \underline{97.95$\pm$0.02} \\
				RNNP & VIT & \underline{98.57$\pm$0.01} & 92.42$\pm$0.05 & 91.22$\pm$0.06 & 90.80$\pm$0.07 \\
				VCIM(our) & VIT &  \textbf{99.17$\pm$0.01} & \textbf{99.10$\pm$0.01} & \textbf{99.07$\pm$0.01} & \textbf{99.04$\pm$0.01} \\
				\hline
			\end{tabular}
		}
		\begin{tablenotes}
			
			\footnotesize
			\item \textit{The results in Tables, \footnotemark[1] by \cite{Liang2022}.
			}
		\end{tablenotes}
	\end{table*}
	
	\begin{table*}[ht]
		\centering
		\caption{Performance of FSL experiments with outlier label noise on TieredImageNet dataset}\label{tab:out2}
		\resizebox{0.5\textwidth}{!}{
			\begin{tabular}{lccccccc}
				\hline
				Method & Backbone & 0\% & 20\% & 40\% & 60\%\\
				\hline
				Matching Networks\footnotemark[1] & Conv4 & 64.99$\pm$0.19 & 60.74$\pm$0.20 & 54.28$\pm$0.21 & 44.93$\pm$0.20 \\
				Vanilla ProtoNet\footnotemark[1] & Conv4 & 71.42$\pm$0.18 & 67.58$\pm$0.19 & 60.97$\pm$0.20 & 50.29$\pm$0.21 \\
				TraNFS-3\footnotemark[1] &Conv4 & 71.13$\pm$0.19 & 67.93$\pm$0.20 & 62.39$\pm$0.22 & 51.82$\pm$0.23 \\
				RNNP\footnotemark[1] & Conv4 & 71.28$\pm$0.18 & 67.29$\pm$0.19 & 60.83$\pm$0.20 & 50.09$\pm$0.21 \\
				Vanilla ProtoNet & VIT & \underline{94.67$\pm$0.06} & \underline{93.96$\pm$0.06} & \underline{93.85$\pm$0.06} & \underline{91.95$\pm$0.08} \\
				RNNP & VIT & 94.42$\pm$0.06 & 84.03$\pm$0.09 & 83.68$\pm$0.09 & 83.58$\pm$0.10 \\
				VCIM(ours) & VIT & \textbf{95.88$\pm$0.05} & \textbf{95.84$\pm$0.06} & \textbf{95.28$\pm$0.07} & \textbf{95.10$\pm$0.07} \\
				\hline
			\end{tabular}
		}
		\begin{tablenotes}
			
			\footnotesize
			\item \textit{The results in Tables, \footnotemark[1] by \cite{Liang2022}.
			}
		\end{tablenotes}
	\end{table*}

	These experimental results show that our model achieves state-of-the-art accuracy with all three types of noise, indicating its superior performance to the other models. It also obtains the lowest variance in all tests, validating its excellent stability in FSL with noisy labels. 
	
	\subsection{Ablation}
	\subsubsection{Clustering methods}
	We used the soft K-means algorithm to generate prototypes for each category in the training. While hard K-means clustering assigns each point to a single cluster, soft K-means allows each point to belong to multiple clusters with a certain probability, offering a more flexible representation of feature distributions. To assess the impact of these two clustering methods on our model's performance, we conduct ablation experiments to compare the contribution of soft K-means and hard K-means to the proposed model in classification.\par
	We conduct tests on the two datasets with 40\% symmetric label swap noise. As shown in Table \ref{tab:kmean}, the soft K-means-based model significantly outperforms the hard K-means-based one. The advantage of soft K-means lies in its ability to assign weights based on the distances between samples and multiple cluster centers. It provides a way to efficiently construct prototypes with little impact by the noisy labels, greatly enhancing the model's accuracy and robustness against label noises.\par
	
	\begin{table}[ht]
		\centering
		\caption{Performance using hard/soft K-means on different datasets with 40\% symmetric label swap noise.
		}\label{tab:kmean}
		\resizebox{0.3\textwidth}{!}{
			\begin{tabular}{lccc}
				\hline
				K-means & MiniImageNet & TieredImageNet \\
				\hline
				Hard & 98.88 & 93.49 \\
				Soft & \textbf{99.11} & \textbf{94.85} \\
				\hline
			\end{tabular}
		}
	\end{table}
	
	\subsubsection{Scale Adaptation}
	The scale adaptation allows the proposed model to adjust the receptive field size, enhancing its ability to adapt to varying input feature distributions. We compare the model performance with the adaptive scale and the fixed scale. As shown in Table \ref{tab:scale}, the accuracy obtained with a fixed scale is significantly lower than the adaptive scale, demonstrating the importance of scale-adaptive algorithms in enhancing model performance. 	
	\begin{table}[ht]
		\centering
		\caption{Performance with and without scale adaptation on different datasets under 40\% symmetric label swap noise.
		}\label{tab:scale}
		\resizebox{0.3\textwidth}{!}{
			\begin{tabular}{lccc}
				\hline
				Method & MiniImageNet & TieredImageNet \\
				\hline
				Fixed Scale & 97.28 & 92.86 \\
				Adaptative Scale & \textbf{99.11} & \textbf{94.85} \\
				\hline
			\end{tabular}
		}
	\end{table}

	\section{Conclusion}
	\label{sec:con}
	We focus on few-shot image classification with noisy labels and propose a robust model VCIM that performs exceptionally well in few-shot learning with noisy labels. The proposed model has an open framework consisting of two fields, embeds a pre-trained deep neural network for feature extraction, utilizes soft K-means clustering to generate prototypes, and employs scale-adaptive techniques to enhance the model's classification accuracy. We test the proposed model with symmetric label swap noise, paired label swap noise, and outlier noise and compare it with state-of-the-art FSL models. The experimental results validate the proposed model achieves superior accuracy in high-noise scenarios, demonstrate its dramatic robustness and stability, and highlight its excellent potential in real-world few-shot learning applications.
	\section*{Acknowledgments}
	This work was supported in part by the Natural Science Foundation of Guangxi under Grant 2025GXNSFAA069486, the National Key R \& D Program of China under Grant 2021YFA1003004, the National Natural Science Foundation of China under Grant 12031003, and the special foundation for Guangxi Ba Gui Scholars.

\printcredits

\bibliographystyle{cas-model2-names}

\bibliography{reference}


\end{document}